# Indoor room Occupancy Counting based on LSTM and Environmental Sensor

Zheyu Zhang
Department of Electrical and Computer
Engineering
Virginia Tech
VA,22043
zheyuzhang21@vt.edu

*Abstract*—This paper realizes the estimation of classroom occupancy by using the $CO_2$ sensor and deep learning technique named Long-Short-Term Memory. As a case of connection with IoT and machine learning, we achieve the model to estimate the people number in the classroom based on the environmental data exported from the $CO_2$ sensor, we also evaluate the performance of the model to show the feasibility to apply our module to the real environment.

## I. INTRODUCTION

Room Occupancy is an important factor in the building system, and has many applications in different areas including building energy management [1], and indoor environment quality [2]. There are many occupancy measurement techniques such as camera [3], Wi-Fi [4], passive infrared sensor (PIR sensor) [5], and so on.

The $CO_2$ sensor is a good kind of IoT device with a low price and a non-intrusive type [6], there are many mature products in the market combined with many environmental sensors, like Netatmo Weather Station.

The weather station is good equipment to measure the indoor environment data, it can detect not only $CO_2$ but also Noise, Humidity, Pressure, and Temperature. Those data are good input data for the machine learning module. The price of the weather station in the market is around $169 including the outdoor station without the outdoor station the price of the product can be lower, so it's a low price of equipment for the users to deploy for a room.

More importantly, the data of the station can be saved on their website and we can export the data at any time with .xls or .cvs format in the data management block on the website, and we can easily edit, add the people number in the period in the same form, this function save our time on the data collection and processing. We also can get real-time data by using Postman not just watching the screen on the weather station and writing down the data. It will be helpful if we can get real-time occupancy data from our module and apply it to other areas, like lighting, and HVAC control to achieve energy savings.

## II. DATA COLLECTION AND PROCESSING

### A. Data Collection

Our experiment place was on the sixth floor of the Virginia Tech Arlington campus building, a classroom 6-053 which can contain around 25 people. Fig. 1 shows we put the weather station in the middle of the classroom since the middle part can have a closer same distance from every seat to increase the accuracy of detection value.

For input data, we can directly download it from the Netatmo weather website, but for output data, the number of people, we have to count by ourselves when the class is beginning. There is a notification function in the Netatmo mobile app that can notify us when the $CO_2$ exceeds a scheduled level, which means there will be people in the classroom, we can go to the classroom when we get the notification and don't have to wait for the whole day since sometime maybe some people come to the classroom for a short meeting.

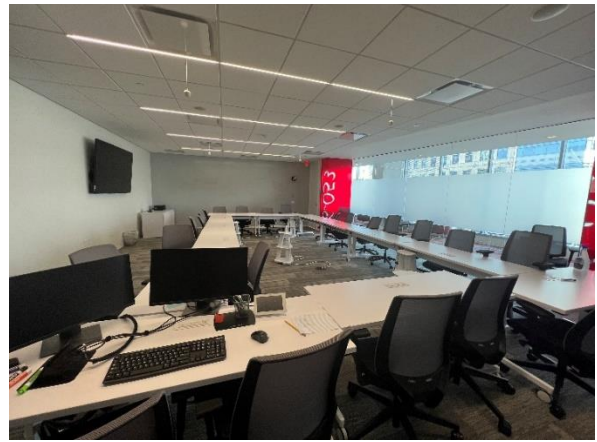

Fig. 1: Weather Station in the classroom

We have collected several days of data for machine learning training that have people in the classroom, since the class schedule limitation, we set seven days as training data, two days as validation data, and two days as test data. Table.1 is an example of the period during the day. The input of the LSTM is always a 3D array, so before loading input data for the LSTM module, we need to reshape the input data into a 3D format. For output data, since we just have one output, we can assume the people number value is categorical data and use one-hot encoding to encode the output data to get better performance.

Table. 1: The data example

| Temp | Hum | CO2 | Noise | Pressure | People |
|---|---|---|---|---|---|
| 21.5 | 43 | 482 | 53 | 1020.8 | 0 |
| 21.6 | 43 | 504 | 52 | 1020.7 | 0 |
| 21.6 | 43 | 497 | 52 | 1020.6 | 0 |
| 21.6 | 43 | 504 | 56 | 1020.6 | 13 |
| 21.7 | 43 | 528 | 59 | 1020.6 | 13 |
| 21.7 | 43 | 535 | 57 | 1020.6 | 13 |

*B. One-hot Encoding*

For the people number before one hot encoding, we should consider that the capacity of the room is around 25 people, but usually, the students in the classroom will not be more than 15. That means the people number higher than 15 is a very low probability event but we cannot ignore it if one day there may be more than 25 people in the classroom for something. So we assume people number more than 15 as 15, which means we put 15 or higher numbers into one category.

One hot is a group of bits of value that only have 0 and 1. The length of the group we set will be 16 which means the category will be from 0 to 15, and for example, the people number is 15 and after one hot encoding in the group of bits, only the index 16th bit will be 1 and all the others will be 0.

After setting the category and one-hot encoding, We will get a good data format for the LSTM model to train. Fig.2 shows the steps of this chapter about the data collection and processing we prepared for the next step.

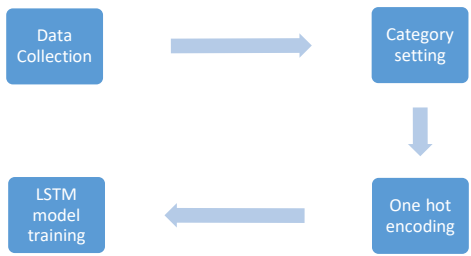

Fig. 2: Data Collection and Processing

### III. LSTM MODULE

*A. Introduction*

Long-Short-Term Memory is a special kind of recurrent neural network that has better performance than traditional recurrent neural networks with many benefits, LSTM can partially solve the vanishing gradient problem or long-term dependence issue.

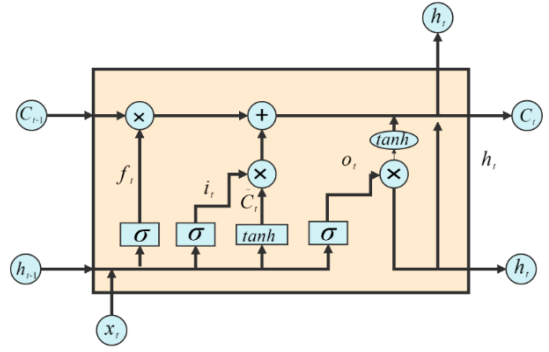

Fig. 3: LSTM cell

Fig.3 is a module from the app named Yitutushi shows the inside architecture of the LSTM cell, the $x_t$ is the input vector to the LSTM unit, $f_t$ is the forget gate's activation, it is the input gate's activation vector, $o_t$ is the output gate's activation vector, $h_t$ is the hidden state vector or output vector of the LSTM unit, $c_t$ is the cell state vector $\tilde{c}_t$ is the cell input activation vector. And the equations of an LSTM cell are below:

$$f_t = \sigma_g(W_f x_t + U_f h_{t-1} + b_f)$$
$$i_t = \sigma_g(W_i x_t + U_i h_{t-1} + b_i)$$
$$o_t = \sigma_g(W_o x_t + U_o h_{t-1} + b_o) \quad (1)$$
$$\tilde{c}_t = \sigma_c(W_c x_t + U_c h_{t-1} + b_c)$$
$$c_t = f_t \odot c_{t-1} + i_t \odot \tilde{c}_t$$
$$h_t = o_t \odot \sigma_h(c_t)$$

The $\sigma_g$ is the sigmoid function $\sigma_c$ and $\sigma_h$ are the hyperbolic tangent function.

*B. Functions setting*

Since we use one-hot encoding to transform the data, the range of loss between the ground truth and the result will be from 0 to 1. So our loss function will be the binary loss function and we also use the Adam optimization algorithm to get a good loss function in order to get a good prediction result. The Adam optimization algorithm is a good algorithm with faster computation time and requires fewer parameters for tuning, so we adapt this algorithm to work as the optimizer in our model.

For the output layer, we know that the output data is one item with one hot encoding format, so we should use the SoftMax activation function to shape the output index range from 0 to 1, and the sum from all one hot index is 1. This will show the probability of the index that shows the probability of the people number generated from the model and then we transform the largest probability index back to the people number, model in this way will show its prediction of the people number based on the input data. Like the Table.2 shows an example of the output data before we transform it into the people number. We can see the thirteenth one is the largest number showing index 12 has the largest probability. So after transforming, the prediction of the people number from our model is 12.

Table. 2: The example of output data before Transforming

| | | |
|---|---|---|
| 3.1594791e-02 | 1.1173296e-03 | 1.9875835e-01 |
| 1.2148099e-01 | 1.7412059e-01 | 9.9778280e-04 |
| 7.6386752e-04 | 3.7641544e-02 | 1.0577185e-03 |
| 4.6020711e-04 | 8.2410325e-04 | 7.5649790e-04 |
| 4.2702064e-01 | 2.0497683e-03 | 9.2915818e-04 |
| 4.2664449e-04 | | |

Above are the LSTM model design. For the next chapter, we use the F1 score to evaluate the model performance.

## IV. RESULT AND EVALUATION

The F1 score is a good measurement combining precision(2) and recall (3) with many benefits that

$$Precision = \frac{True\ Posistives}{True\ Positives + False\ Positives} \quad (2)$$

$$Recall = \frac{True\ Posistives}{True\ Positives + False\ Negatives} \quad (3)$$

Small precision or recall will lower the overall score, so it (4) has a good balance between the two metrics

$$F1\ Score = \frac{2 * Precision * Recall}{Precision + Recall} \quad (4)$$

The higher F1 score means a higher precision and recall of the model. We can simply assume a higher F1 score can show better performance of our model. We will compare two models one applies one hot encoding and the other one without one hot encoding.

### A. Performance Comparison

We choose two days value as the test data to make a prediction and we get predictions by using our model, we first build up the model without one-hot encoding for the output data, then we extend our model with one-hot encoding as an optimization, and the performance clearly shows the essential to apply one-hot encoding.

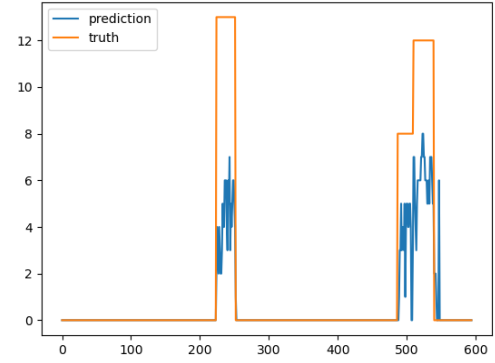

Fig. 4: Result of the model without one-hot encoding

In Fig.4 and 5, the orange line is the real number of people that we count, and the blue line is the predicted number of people generated by the model

Fig.4 shows the performance without one-hot encoding and the input data and output of data of Fig. 4 and 5 are the same. So it's convincing to make a comparison between these two figures. Fig.4 shows the people number varies and trends but the people number can not be 1.5 or 10.1 like other data such as price, score, and weight, so there is not good to show the tendency of people number variation. However, after one-hot encoding, the people number is closer to the ground truth value, and the prediction shows the exact number of people. The F1 score also shows the comparison of two figures that 0.8 for non-one-hot and 0.85 for the one-hot result. As a result, we can see that one hot encoding has better performance in that it can show the exact number of people and it can increase the precision and recall metrics for the model.

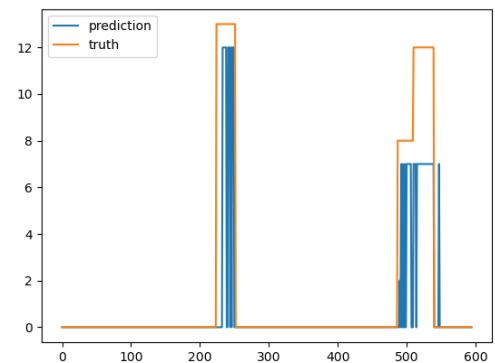

Fig. 5: Result of the model with one-hot encoding

*B. Limitations*

For our model, we find some limitations but can improve in the future.

First, the data size is small which bring our F1 score result cannot exceed 0.9 (very good). Because preliminary work costs a lot of time like, $CO_2$ station setup, the permission of placement in the classroom, and deployment. We also make some mistakes during collecting data, like just counting people's numbers but forget writing down the time for that period, and sometimes there's a sudden small meeting out of schedule happening at that classroom, and we cannot arrive at the place to count immediately.

Second, since the CO2 data has a 5-minutes data update delay, the real-time property may face challenges to achieve.

Third, about data collection, we can only count the constant people number in the classroom by hand, if some people are going outside to the restroom for ten minutes we cannot notice because it is hard to stay in the classroom to count the dynamic number. So, we need a better method to count people's numbers to increase the accuracy of the data.

The last one is because the HAVC engages in the classroom to exchange the air from room to outside, the $CO_2$ value may be affected by HVAC power, if the power of HVAC is at a high period, $CO_2$ value will decrease significantly when people left. In contrast, when HAVC is not working like on Sunday, the $CO_2$, temperature value will stay at a high level that is different from the weekdays. In our model, we ignore the weekend value because there are no people in the classroom, but when we apply this model in the classroom, we cannot ignore the weekend value. We will try to figure out this problem on the application.

## V. Future Work

Occupancy data is important for its application. Like we can control the light and HVAC based on the occupancy in the room to save energy. Our work will focus on improving our model performance and exploring its application.

We will try to collect more data and improve our collection method, like using the camera in the classroom to count people in the different time stamps, and we also can collect data next semester since the class this semester is going to end to enlarge the size of the dataset.

The application of the model will be more attractive for us to do in the future after we get good performance for our model. We are going to apply our model to a Building Energy Management System (BEMS). We find a platform named Building Energy Management Open-Source Software (BEMOSS) [7] that has the potential ability to apply our model to lighting control or HVAC control. We can change the brightness of the light and schedule the HVAC working time via BEMOSS, combined with our model, we want to achieve auto-control for the light and HVAC in the classroom, for example, the model can detect the occupancy in the classroom, and BEMOSS control the lightness or HVAC power to face the demand of the classroom. With more people in the room, we can increase the lightness and set HVAC to set a comfortable environment for students in the classroom during the study, when there are no people or too few people, there is no reason to turn on all lights in the room with 100 percent power.

Above are our work on occupancy estimation based on the LSTM.